\documentclass[conference]{IEEEtran}
\IEEEoverridecommandlockouts

\usepackage{cite}
\usepackage{amsmath,amssymb,amsfonts}
\usepackage{algorithmic}
\usepackage{graphicx}
\usepackage{textcomp}
\usepackage{xcolor}
\def\BibTeX{{\rm B\kern-.05em{\sc i\kern-.025em b}\kern-.08em
    T\kern-.1667em\lower.7ex\hbox{E}\kern-.125emX}}
\begin{document}

\title{Agentic Network Traffic Monitoring\\
\thanks{Research was sponsored by the Department of the Air Force Artificial Intelligence Accelerator and was accomplished under Cooperative Agreement Number FA8750-19-2-1000. The views and conclusions contained in this document are those of the authors and should not be interpreted as representing the official policies, either expressed or implied, of the Department of the Air Force or the U.S. Government. The U.S. Government is authorized to reproduce and distribute reprints for Government purposes notwithstanding any copyright notation herein.}
}

\author{\IEEEauthorblockN{
Manuel Tsoukatos$^{1,2}$, Hayden Jananthan$^{2}$, Jeremy Kepner$^{2}$
\\
\IEEEauthorblockA{
$^1$United States Air Force \\
$^2$Massachusetts Institute of Technology
}}}

\maketitle

\begin{abstract}
As the use of agentic artificial intelligence increases in nearly every industry, there exists a widening attack surface. It is necessary to monitor agents to ensure that agents are acting in a way that is aligned with the users intent. Auditing an agent's network traffic provides a clear record of the agent interactions. This work presents a novel approach to monitoring the network traffic of agentic systems using complex valued hypersparse traffic matrices by integrating DBOS (DataBase OS), the OneSparse PostgreSQL database, and the GraphBLAS math library.  To develop these concepts an agentic simulator was constructed, allowing a varying numbers of AI agents to collectively survey a virtual environment using different strategies. The resulting network traffic matrices enable easy monitoring of the AI agents.
\end{abstract}

\section{Introduction}
Artificial intelligence is an incredibly fast growing field and over the last year the rise in AI agents has accelerated. The use of AI agents  in web services, financial applications, and many other applications, requires that users have ways to audit and monitor agent behavior. \cite{HOSSEINI2025100399}. AI agents have potential to significantly enhance efficiency, but at the same time create significant security vulnerabilities with things like prompt injections, data leaks, and adversarial actors \cite{agentic_risks_Su, Murugesan_2025, Deng_2025}.

Network traffic is at the heart of agentic systems. As the data that flows across a network from source to destination it may cross many other nodes while in transit, and the modeling/analysis of network traffic can provide insights into agent behavior \cite{Jones_2023, Lakhina_2004}.

Traffic matrices have emerged as compact, efficient, and scalable way to monitor network traffic. A common  use of traffic matrices defines rows as sources and the columns as destination. The value in a particular (row, column) position corresponds to the number of packets of information sent from that source to that destination \cite{Kepner_2019}.  If the sources and destinations span a large potential range (e.g., network Internet Protocol - IP - address) they can be very large. Because the vast majority of values in the matrix are 0 it is common to use hypersparse matrix to store these matrices. GraphBLAS is an open source standard library for creating and operating on hypersparse matrices \cite{Davis-Graphblas-2019, Trigg_2022, Jones_2022}. More specifically for this project, the PostgreSQL extension OneSparse,  binds the GraphBLAS library to PostgreSQL databases and enables GraphBLAS matrices to be operated on as a PostreSQL type with an SQL table \cite{pelletier2024onesparse}.

DBOS (Database-Oriented Operating System) is a durable workflow system that has become popular for supporting long running AI agents.  DBOS connects to PostreSQL database to save all events as structured data in a database \cite{Skiadopoulos_2021}. Recent DBOS advancements have added workflow guarantees that ensure: exactly once execution with no workflow duplication, every workflow runs to completion even if a crash occurs, and logically correct and reverse workflow actions 
\cite{li2025dbos}.
Because agentic workflows are often complex and run for potentially very long times, DBOS and its guarantees, provide an excellent way to track and manage AI agent workflows, by saving state and allowing for durable execution \cite{stonebraker2026consistency}.

Combining of these different technologies (DBOS, Graphblas, Onesparse) can be used to create a scalable robust system for tracking and analyzing agentic workflows and their networ traffic \cite{lockton-2025, lockton2025dbos}.  Our work applies these concepts in the context of an agentic simulator that allows varying numbers of AI agents to collectively survey a virtual environment using different strategies. The resulting network traffic matrices demonstrate the ability to monitor AI agents with this approach that requires no internal knowledge of the agents.

\section{Method}

The approach taken is a customizable testbed to simulate multi-agent interaction. The goal is to try and encompass as many different scenarios as possible, in physical situations where range/distance factors may apply as well as other  variations. To begin, a world map is generated. It is a two-dimensional grid of specifiable variable size, with a combination of impassable walls, and passable open space. There is an orchestrator agent, fixed in place in the top left corner and its goal is to create a mental map of the ground truth environment using subagents that can move around the environment.

\begin{figure}
    \centering
    \includegraphics[width=\linewidth]{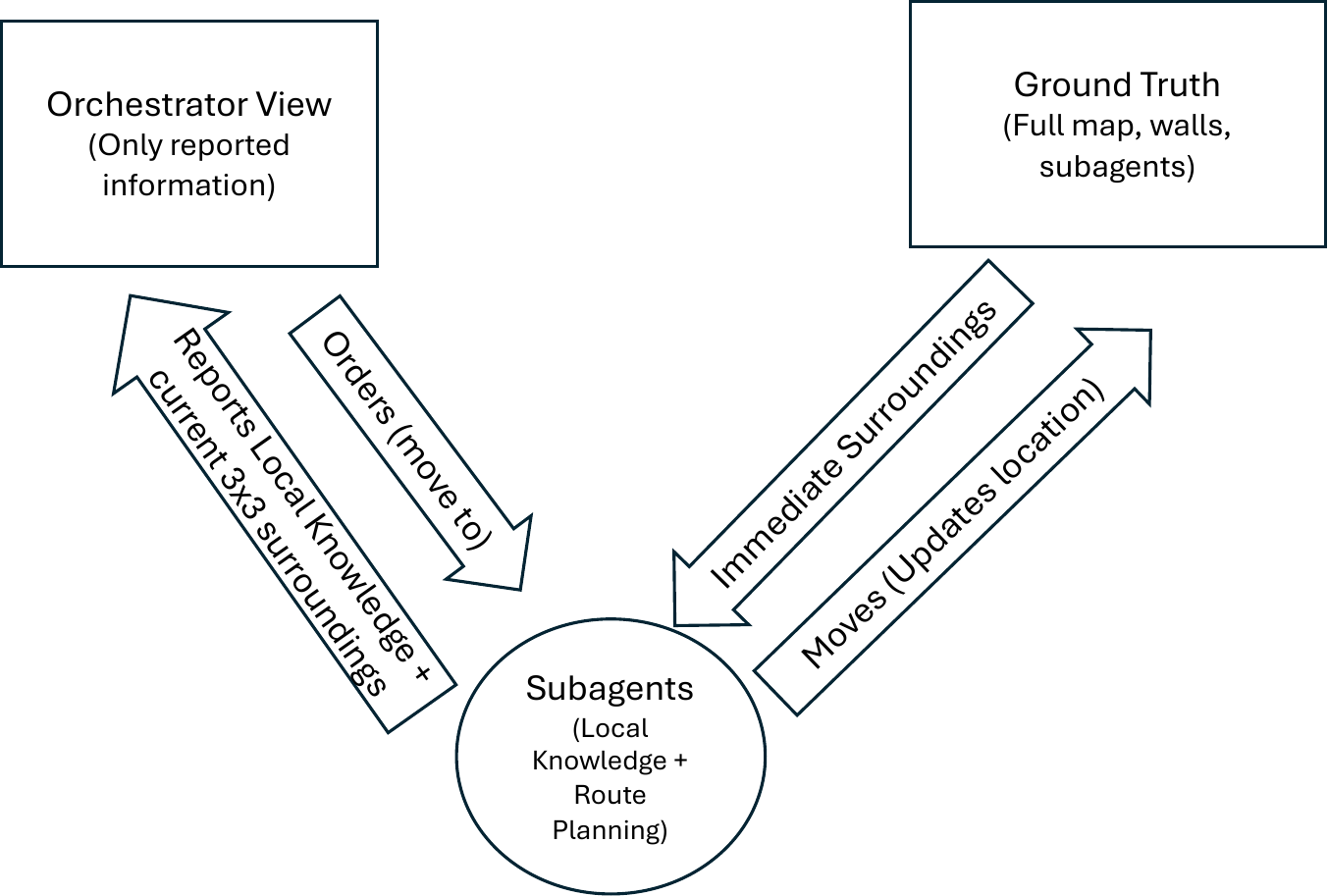}
    \caption{Information flow diagram}
    \label{fig:information}
\end{figure}

The orchestrator and subagents have different views and actions they can take. The orchestrator only sees information that is relayed back to it from what the subagents directly observe in their immediate three by three grid surroundings. As can be seen in Figure \ref{fig:information} the orchestrator passes orders to the subagents which tell them to go to an unkown position. Subagents, upon moving, interact with the ground truth database with DBOS steps that write the subagents new position, and read the new surroundings. The subagents maintain local knowledge of all places  they have been and do route planning based off of this knowledge. After an agent moves and observes something new, it reports back what it has have seen. When arriving at assigned destination, the agent  sends a ready message to the orchestrator who then provides a new destination.

\begin{figure}
    \centering
    \includegraphics[width=\linewidth]{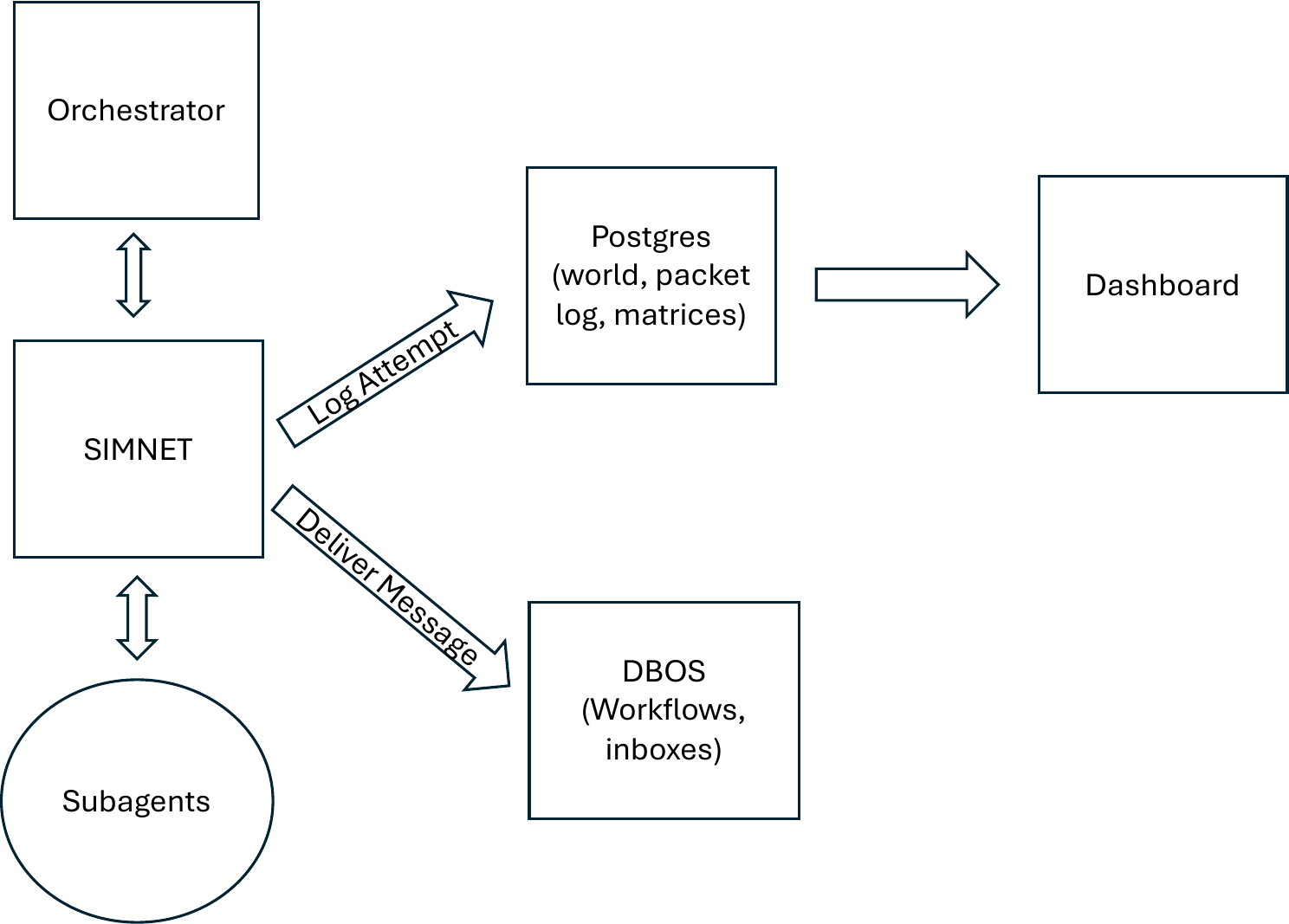}
    \caption{Communication protocol diagram}
    \label{fig:comms}
\end{figure}

The only way the 
orchestrator can talk to the subagents is through a custom function that simulates network communication. In Figure \ref{fig:comms} this is visible with the SIMNET block. This simulated networking produces network traffic logs that are then stored in a PostgreSQL Database. If the message is able to be completed by meeting the connection success criteria, it is sent to the recipient with the DBOS.send command that durably logs the message. The result of whether the message was successfully delivered is also logged with the DBOS.step command to store all traffic for later analysis. This is then displayed in dashboard for live monitoring and for visualization.

The network traffic, is then aggregated into two complex valued GraphBLAS matrices using OneSparse. One matrix holds the true source to true destination of every message. A second maps all intermediary connections or hops along the route, which will differ from the first matrix if the message was passed to other recipients before the final destination.  Agentic systems engaging in very dynamic communications over a range of time scales and frequently timeout or abort activities. Thus, a key element to track in agentic systems is unsuccessful messages sent. For both matrices, the real component is the number of successfully delivered messages and the imaginary component is the number of unsuccessful messages sent.   In both cases, the entries of a complex valued traffic matrix A are given by the following

\[{\bf A}(i,j) = v + u \sqrt{{\scriptstyle -}1}\]

where v is the number of successful communications sent from i to j and u is the number of unsuccessful communications attempted from i to j. The total number of messages sent is given by the matrix $Re({\bf A}) + Im({\bf A})$.  This approaches takes advantage of the builtin GraphBLAS support for complex values and is an efficient way to track a more complete picture of the network traffic. In the simulation display, only the real component is shown unless the imaginary dropped component is non-zero, in which case it is displayed as (real, imaginary). 

There are many potential agent control scenarios that can be explored with this simple system. The baseline scenario involves a simple heuristic where the orchestrator agents select the nearest unknown position and send to the nearest  subagent ready to receive an order. The baseline scenario assumes the orchestrator has constant two way communication with the subagents with unlimited  range. The subagents send reports to the orchestrator to convey information and to request a new order. When subagents receive a new order, they perform a breadth-first search (BFS) over their local knowledge of places they have seen.

The baseline scenario creates many possibilities for variation that includes
\begin{enumerate}
\item Subagents sending peers messages with updates about their local knowledge
\item A range limitation which prevents messages from being delivered if the distance between two parties is greater than a specific value
\item Daisy chaining as a way to solve the range limitation problem where subagents can act as a mesh network and pass messages intended for other recipients to the next closest subagent or true recipient
\item Delegation that creates a hierarchy of subagents where higher ranked subagents receive general areas to explore and then divide it up amongst themselves and their subordinate subagents
\item Using an large language model (LLM) as the brain for the reasoning instead of a simple heuristic
\end{enumerate}

\section{Results}

\begin{figure*}[!h]
    \centering
    \includegraphics[width=\linewidth]{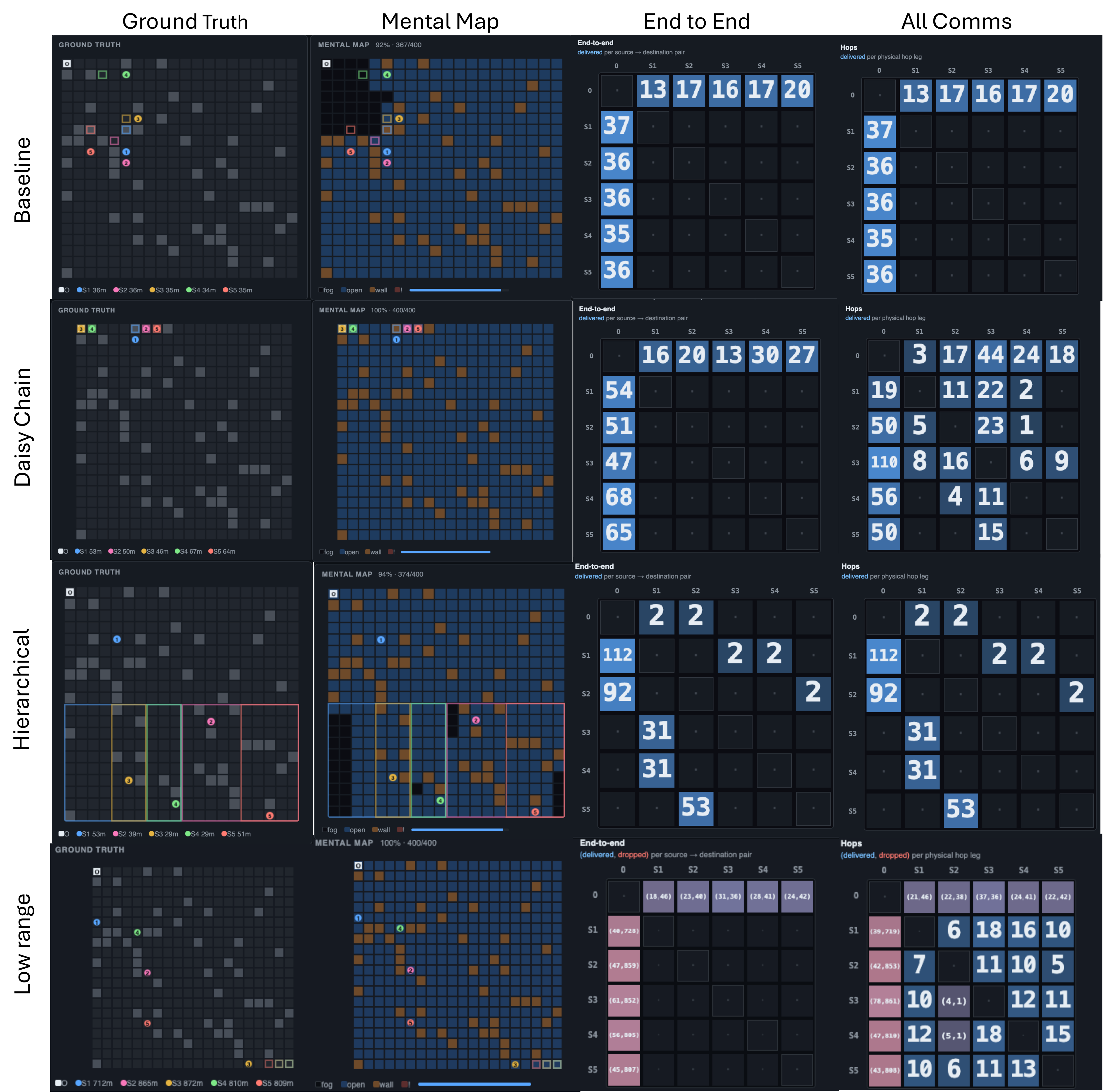}
    \caption{The above displays the key aspects of each simulation. The rows are individual simulations, showcasing the baseline, daisy chain, hierarchical, and low-range daisy chain scenarios. The columns are the two maps (ground truth and mental map), and two traffic matrices (true source/destination pairs and all intermediary communication hops). The maps and everything besides the communication protocols were kept the same to compare the resulting traffic matrices. }
    \label{fig:All_results}
\end{figure*}

All of these different settings and design choices allow for a wide variety of robust situations to simulate agentic behavior. The traffic matrices provide a way to visualize the often invisible connections between agents as they move autonomously in any task. This testbed serves as a foundation for monitoring agentic behavior for enabling detecting of anomalous and adversarial behavior.

Figure \ref{fig:All_results} illustrates four different scenarios. The first row demonstrates the baseline scenario with an unlimited range policy. In this snapshot the dashboard on the left most map shows the ground truth environment, where dark squares represent a free open space and light squares represent an open space. To its right is the mental map reconstruction at 92\% completion to show what  progress within the simulation looks like. The highlighted squares are the target of the respectively colored subagent given by the colored circles. As is expected, the two traffic matrices showing true source/destination pairs looks identical to the all hops matrix, because there is no intermediary communication. This is what a standard operational technology (OT) network would look like where one source goes to all destinations, and all destinations  go back to that source. In this scenario an anomaly can be  easy to spot because all other positions besides the first row and column should be zero. 

In the daisy chaining scenario there is a manhattan distance range limit of 20 squares. This scenario has run to completion and the shutdown order to end the simulation was given by the orchestrator after the mental map reached 100\%. The end to end matrix has the same shape as the baseline which is expected because all messages are between the orchestrator and subagents. However, the intermediary matrix diverges because many messages were routed through closer subagents. In this example, the subagents maintain consistent connectivity and communication throughout the run. This is ideal for an agentic swarm as no time is wasted with dropped packets and subagents can continue exploring without also making moves to reconnect in range.

In the hierarchical example in the third row, the orchestrator gives more general orders to its direct subordinates. The chain of command in this scenario is the orchestrator to the blue (one) and magenta (two). Blue (one) is in charge of yellow (three) and green (four). Magenta (two) is in charge of pink (five). Blue and Magenta receive orders to explore the left and right regions, respectively, which they divide up among themselves and subordinate subagents. This is visible with the colored highlighted regions which show the territory area of responsibility. Because there is no range limit here, the end to end and hops matrices are identical to each other, but because of the hierarchy are very different from the previous scenarios. Subagents one and two are the only ones interfacing with the orchestrator, collecting updates from subordinates and aggregating into updates for the orchestrator. The higher ranked subagents communicate with their subordinates in the matrix, and a very clear picture of the chain of command is visible here. This is very representative of how many AI agents  spin up subagents to accomplish intermediary tasks and in this view would make any deviation from the chain of command apparent.

In the final example in the fourth row, daisy chaining with a very low range of 5 squares was introduced. This essentially forced every subagent to return to within 5 squares of the orchestrator to relay the message of new findings and request new orders. This produces a significant amount of dropped packets as subagents attempt communication and are often times unable to deliver the packet even with a chain of them, due to a lack of coordinating. This does provide a useful view on the specific number (dropped packets)  that could be minimized when exploring new ways to maintain connectivity in more challenging situations.

\section{Conclusion and Future Work}

Using a combination of DBOS and PostgreSQL to preserve the state and manage agentic workflows, and GraphBLAS to model network traffic, there is now a customizable simulation to explore agentic communication patterns. There are a multitude of existing scenarios and configurable settings, but in the future, this involves increasing the library of things to simulate. Running experiments with this test bed at a scale of 1000s of runs across 1000s of subagents could provide useful data on agentic behavior. The system could also be used to create an artificial datasets for both normal and anomolous agentic network patterns. Future work may also entail adding this type of network monitoring into current agentic systems, as a way of auditing all communications between agents. Additionally, one could more efficiently store network traffic as the phase between delivered and dropped packets, resulting in a real matrix that holds more information than a traditional network traffic matrix with just delivered packets logged.

\section*{Acknowledgment}
The authors wish to acknowledge the following individuals for their contributions and support 
L. Anderson, W. Arcand, D. Bestor, W. Bergeron, B. Bond, C. Byun, A. Bonn, D. Burrill, V. Gadepally, J. Gottschalk, T. Hardjono, M. Houle, M. Hubbell, M. Jones, P. Luszczek, P. Michaleas, L. Milechin, J. Mullen, C. Leiserson, C. Milner, S. Mohindra,  A. Pentland, A. Prout, C. Prothmann, A. Reuther, A. Rosa, D. Rus, D. Ross, J. Ross, M. Sherman, S. Van Broekhoven, M. Weems, J. Wilkinson, C. Yee.

\bibliographystyle{ieeetr}
\bibliography{AgenticTraffic}

\end{document}